# Data-Driven Abdominal Phenotypes of Type 2 Diabetes in Lean, Overweight, and Obese Cohorts

Lucas W. Remedios[a]*, Chloe Cho[e], Trent M. Schwartz[b], Dingjie Su[a], Gaurav Rudravaram[b], Chenyu Gao[b], Aravind R. Krishnan[b], Adam M. Saunders[b], Michael E. Kim[a], Shunxing Bao[b], Alvin C. Powers[j,k,l], Bennett A. Landman[a,b,e], John Virostko[f,g,h,i]

[a]Vanderbilt University, Department of Computer Science, Nashville, USA; [b]Vanderbilt University, Department of Electrical and Computer Engineering, Nashville, USA; [e]Vanderbilt University, Department of Biomedical Engineering, Nashville, USA; [f]University of Texas at Austin, Department of Diagnostic Medicine, Dell Medical School, Austin, USA; [g]University of Texas at Austin, Livestrong Cancer Institutes, Dell Medical School, Austin, USA; [h]University of Texas at Austin, Austin, Department of Oncology, Dell Medical School, USA; [i]University of Texas at Austin, Oden Institute for Computational Engineering and Sciences, Austin, USA; [j]Vanderbilt University Medical Center, Department of Medicine, Division of Diabetes, Endocrinology, and Metabolism, Nashville, USA; [k]Vanderbilt University, Department of Molecular Physiology and Biophysics, Nashville, USA; [l]VA Tennessee Valley Healthcare System, Nashville, USA

**Abstract**

**Purpose:** Although elevated body-mass index (BMI) is a well-known risk factor for type 2 diabetes, the disease's presence in some lean adults and absence in others with obesity suggests that more detailed measurements of body composition may uncover abdominal phenotypes of type 2 diabetes. With artificial intelligence (AI), we can now extract detailed measurements of size, shape, and fat content from abdominal structures in 3D clinical imaging at scale. This creates an opportunity to empirically define body composition signatures linked to type 2 diabetes risk and protection using large-scale clinical data. **Approach:** We studied imaging records of 1,728 de-identified patients from Vanderbilt University Medical Center with BMI collected from the electronic health record. To uncover BMI-specific diabetic abdominal patterns from clinical computed tomography (CT), we applied our design four times: once on the full cohort (n = 1,728) and once on lean (n = 497), overweight (n = 611), and obese (n = 620) subgroups separately. Briefly, our experimental design transforms abdominal scans into collections of explainable measurements, identifies which measurements most strongly predict type 2 diabetes and how they contribute to risk or protection, groups scans by shared model decision patterns, and links those decision patterns back to interpretable abdominal phenotypes in the original explainable measurement space of the abdomen using the following steps. 1) To capture abdominal composition: we represented each scan as a collection of 88 automatically extracted measurements of the size, shape, and fat content of abdominal structures using TotalSegmentator. 2) To learn key predictors: we trained a 10-fold cross-validated random-forest classifier with SHapley Additive exPlanations (SHAP) analysis to rank features and estimate their risk-versus-protective effects for type 2 diabetes. 3) To validate individual effects: for the 20 highest-ranked features, we ran univariate logistic regressions to quantify their independent associations with type 2 diabetes. 4) To identify decision-making patterns: we embedded the top-20 SHAP profiles with Uniform Manifold Approximation and Projection (UMAP) and applied silhouette-guided K-means to cluster the random forest's decision space. 5) To link decisions to abdominal phenotypes: we fit one-versus-rest classifiers on the original anatomical measurements from each decision cluster and applied a second SHAP analysis to explore whether the random forest's logic had identified abdominal phenotypes. **Results:** Across the full, lean, overweight, and obese cohorts, the random-forest classifier achieved mean area under the receiver operating characteristic curve (AUC) of 0.72–0.74. SHAP highlighted shared type 2 diabetes signatures in each group—fatty skeletal muscle, older age, greater visceral and subcutaneous fat, and a smaller or fat-laden pancreas. Univariate logistic regression confirmed the direction of 14–18 of the top 20 predictors within each subgroup ($p < 0.05$). Clustering the model's decision space further revealed type 2 diabetes-enriched abdominal phenotypes within the lean, overweight, and obese subgroups. **Conclusions:** We found similar abdominal signatures of type 2 diabetes across the separate lean, overweight, and obese groups, which suggests that the abdominal drivers of type 2 diabetes may be consistent across weight classes. Although our model had a modest AUC, the explainable components allowed for clear interpretation of feature importance. Additionally, in both lean and obese subgroups, the most important feature for identifying type 2 diabetes was fatty skeletal muscle.

*Lucas W. Remedios**, E-mail: lucas.w.remedios@vanderbilt.edu

# 1 Introduction

Type 2 diabetes is a chronic metabolic disease linked to alterations in body composition[1–5]. While coarse measurements like body mass index (BMI) and waist circumference are correlated with type 2 diabetes, the pattern of fat distribution is a stronger predictor of disease risk than overall adiposity[6–8]. Importantly, people with similar BMI can exhibit vastly different metabolic risk profiles. Some obese individuals remain type 2 diabetes-free (metabolically healthy obesity), while others with normal BMI develop type 2 diabetes.

In the type 2 diabetes population, changes have been noted across many anatomical structures. When compared to normal populations, there is evidence for higher body fat percent, more subcutaneous fat, more visceral fat, as well as less and fatty lean mass like skeletal muscle[9,10,11,12,13,14,15,16,17]. While BMI defines obesity, closer examination has shown that in obese adults an increase in visceral fat is associated with type 2 diabetes[15,18]. In the abdominal organs, type 2 diabetes is associated with an increase in fat in the liver, pancreas, and kidneys, as well as a smaller pancreas[19,20,21,21,22,23,24,25–27,15,23,28]. Since low muscle mass is related to type 2 diabetes, it follows that resistance training has been shown to be associated with an improvement in blood sugar and insulin sensitivity[29]. However, there is some conflicting evidence from bioelectrical impedance analysis implying that lean mass is not protective of type 2 diabetes[30,31].

Today, medical imaging and artificial intelligence make it possible to measure abdominal composition with unprecedented detail, scale, and interpretability[32–36]. Tools such as TotalSegmentator automatically segment over 100 anatomical structures in CT or MRI scans, enabling large-scale extraction of organ volumes, fat distributions, and shape metrics without manual intervention[37,38]. When paired with explainable AI techniques such as SHAP (SHapley Additive exPlanations), predictive models built on these features can rank the importance of each anatomical measurement and quantify its directional influence on type 2 diabetes risk[39]. This explainability transforms AI from a black-box into a tool for scientific discovery, both validating known relationships (e.g., high visceral fat increases risk) and revealing novel combinations that may not emerge from a purely hypothesis-driven analysis. In this context, data-driven modeling becomes a powerful means to help uncover new phenotypes and generate mechanistic validation of the literature.

While considerable effort has been devoted to phenotyping specific organ systems and structures in type 2 diabetes, our work focuses on characterizing abdominal signatures of type 2 diabetes across separate lean, overweight, and obese groups. We use a holistic view of the abdomen to assess combinations of features to 1) rank feature importance for type 2 diabetes prediction, 2) discover multi-feature signatures of type 2 diabetes, and 3) explore the model's decision-making landscape to investigate anatomical phenotypes (Figure 1).

## 2 Methods

We take heterogenous clinical imaging data, represent each scan as a collection of abdominal measurements, and determine which patterns of features separate type 2 diabetes from control (Figure 2). The computational system we have designed in this work connects validated tools into a pipeline for pattern discovery.

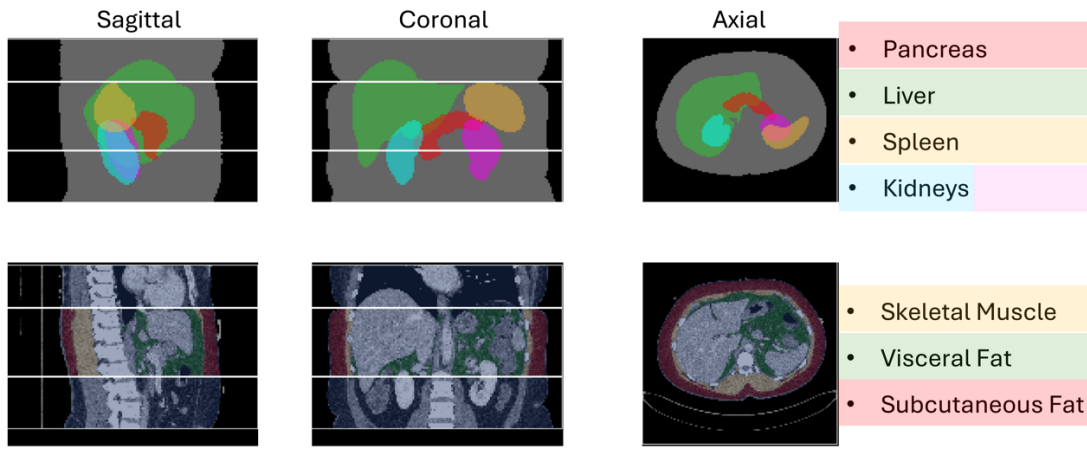

**Fig. 1** AI-driven body composition analysis of large imaging datasets opens new avenues for data-driven discovery of abdominal phenotypes linked to metabolic risk, particularly by identifying which quantitative measurements, and which combinations of those measurements, most effectively distinguish individuals with type 2 diabetes from control patients.

*2.1 Data*

Leveraging a culture of secondary data use[40], we conducted a retrospective observational study on de-identified data from Vanderbilt University Medical Center under Institutional Review Board approval (IRB # 241494). After processing and quality assurance, the dataset consisted of 1,728 patients, aged 20 or older, each with a single clinically acquired CT scan of the abdomen.

*2.2 Automatic Measurement of the Abdomen and Quality Assurance*

Each of the 1,728 scans successfully passed through the processing steps and quality assurance procedure detailed in the rest of this section. The medical images were converted from DICOM[41] to NiFTI[42] format using dcm2niix[43]. Scans were reoriented to standard orientation (LAS), manually inspected with AutoQA[44], cropped between the L5 and T7 vertebrae (from TotalSegmentator vertebrae segmentation[45]) and resampled to 3 mm isotropic resolution to speed up downstream processing steps.

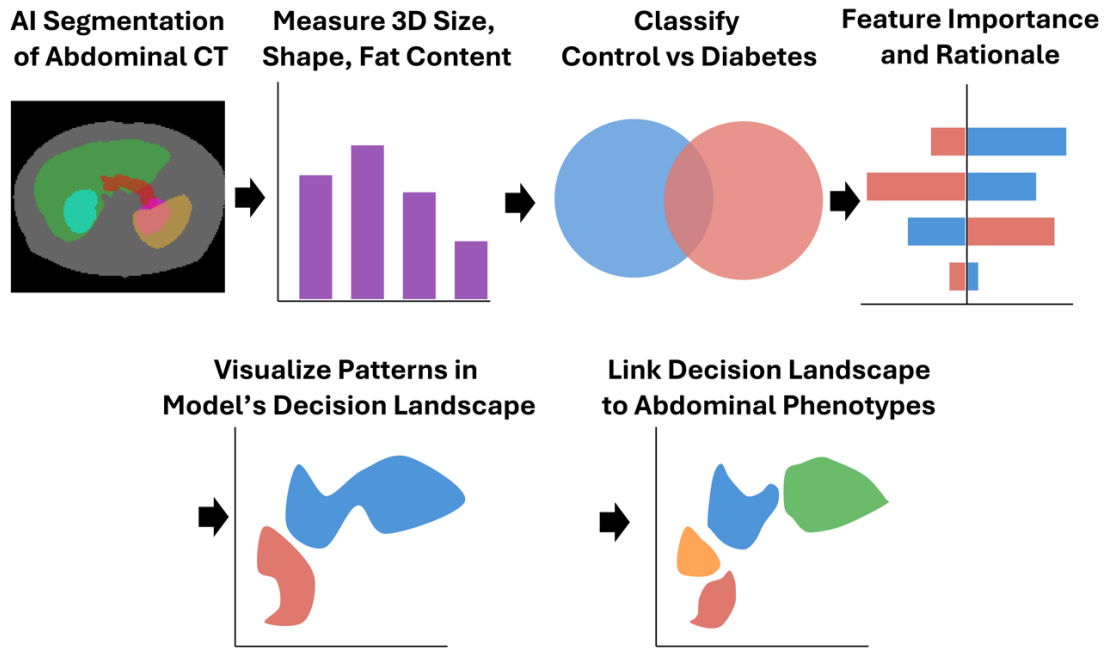

**Fig. 2** Our approach for abdominal phenotyping leverages AI to represent the abdomen as a combination of 88 fully explainable 3D measurements, capturing the size, shape, and fat content of abdominal structures. We use predictive modeling with explainable AI techniques to characterize differences in the abdominal anatomy of individuals with type 2 diabetes. By mapping decisions from the predictive models back to anatomy, our pipeline functions both as a pattern-discovery engine, revealing emergent phenotypes, and a validation engine, highlighting where data-driven insights align with existing knowledge.

These preprocessed data were then fed to the CT model from TotalSegmentator[37] version 2.8 for automated segmentation of the pancreas, liver, spleen, right kidney, and left kidney. Additionally, these data were segmented for visceral fat, subcutaneous fat, and skeletal muscle in an inferior-to-superior field of view between the middle of the L2 to T10 vertebrae. We created an explainable representation of the abdomen with 88 measurements by taking the 8 segmentations, which delineate the abdominal structures in 3D, and extracting 11 size, shape, and CT Hounsfield unit density measurements: volume, surface area, surface area to volume ratio, elongation, flatness, sphericity, major axis length, least axis length, minor axis length, maximum 3D diameter, and median intensity (median Hounsfield units on CT) with PyRadiomics[46] version 3.1. Median intensity of soft tissue on CT inversely correlates with fat content[47]. The scans in our dataset passed an extensive quality control pipeline, including the automatic exclusion of edge-touching segmentations (structures leaving the image field of view), and multi-island segmentations (more than one recognized object for structures that are one object), as well as several rounds of careful manual visual review of both images and segmentations to ensure high data quality and reliable measurements. Type 2 diabetes status was assigned from ICD/PhecodeX[48] records up to one year after imaging while excluding cases with cancer, pancreatic disease, sepsis, or trauma. CTs were automatically categorized as contrast-enhanced or non-contrast via TotalSegmentator's phase classifier and confirmed by visual inspection with the high-throughput AutoQA quality assurance tool[44].

*2.3 BMI-Based Subgrouping and Data Balancing*

To investigate if abdominal patterns change in different BMI classes, we grouped our data into four cohorts: All BMI (all the data), lean (BMI < 25 kg/m$^2$), overweight (25 kg/m$^2$ ≤ BMI < 30 kg/m$^2$), and obese (BMI ≥ 30 kg/m$^2$). These BMI groupings were selected following guidance from the World Health Organization. We present the data and demonstrate that the distributions were reasonably balanced across sex, BMI, and age (Figure 3).

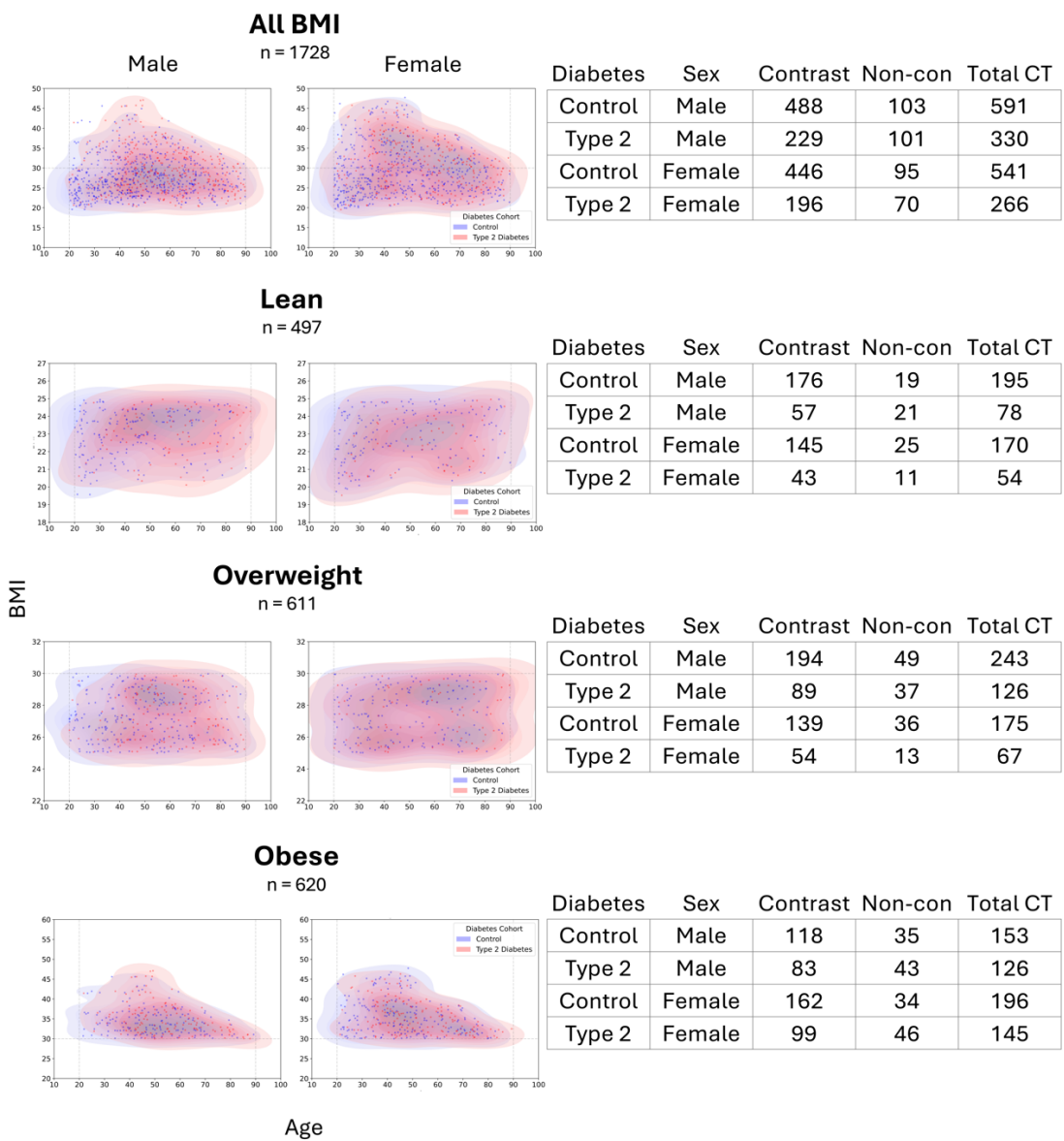

**Fig. 3** Our balanced data supports answering key scientific questions such as 1) why some lean individuals develop type 2 diabetes; 2) why some obese individuals remain type 2 diabetes-free. We stratified a large dataset of patients with clinical CT of the abdomen into four groups: the full cohort (All BMI), lean (BMI < 25 kg/m$^2$), overweight (25 kg/m$^2$ ≤ BMI < 30 kg/m$^2$), and obese (BMI ≥ 30 kg/m$^2$). Within each subgroup, the BMI distributions over the age range were reasonably comparable between individuals with and without type 2 diabetes, visualized separately for male and female.

*2.4 Assessing Separability, Feature Ranking, and Signatures of Type 2 Diabetes*

To investigate separability of control from type 2 diabetes based on our collection of abdominal measurements, we trained a random forest classifier under 10-fold cross-validation. Our cross-validation data splits were stratified on the combination of type 2 diabetes label, sex, age in decade, and whether the CT had contrast. The inputs to the random forest were the 88 abdominal size, shape and intensity features, along with confounders: sex, age, BMI, and CT contrast (contrast or non-contrast). The output of the model was an estimated probability of type 2 diabetes (0 mapped to control, 1 mapped to type 2 diabetes). The random forest used 300 estimators, a max depth of 10, and specified that a leaf node needed at least five examples. To address class imbalance on the type 2 diabetes label, we used balanced class weighting from scikit learn's random forest class weight option during training[49]. Performance was measured via AUC, which was computed separately for each fold and averaged across folds. Each validation fold where performance was measured had ~10% of the total data.

To explain why the random forest made decisions, we used SHAP analysis. We concatenated the SHAP values across the 10 folds and then identified the top 20 most important features for the random forest to decide if our scans were from control or type 2 diabetes patients.

*2.5 Independent Significance Testing of Model-Prioritized Features*

To verify that the random forest's SHAP-ranked features were independently informative, we evaluated each of the top 20 measurements in a separate (univariate) logistic regression model, while adjusting for key confounders—sex, age, BMI, and a binary CT-contrast indicator. These logistic regressions were fit on the entire dataset (no cross-validation) for each of the four cohorts separately. For every model we calculated the odds ratio (OR), 95 % confidence interval, and standard error. The *p*-values were corrected for multiple comparisons with the Benjamini-Hochberg false discovery rate (FDR) across the 20 tests in each BMI stratum, and features with adjusted $p < 0.05$ were deemed statistically significant after FDR correction. Finally, we compared the sign of each OR with the SHAP directionality to measure agreement between the multivariate random-forest findings and these confounder-controlled univariate analyses.

*2.6 Measuring Emergent Phenotypes*

To determine whether the random-forest decision space contained distinct anatomical phenotypes, we first converted each scan's SHAP[39] profile (20 features) into a two-dimensional embedding with UMAP[50,51]. We then applied k-means[51,52] clustering to the UMAP coordinates, scanning k = 2–10 and automatically selecting the solution with the highest silhouette[52] score.

For every resulting cluster, we trained a one-vs-rest random forest classifier using the abdominal measurements and confounders (sex, age, BMI, CT-contrast flag) to quantify how its anatomy differed from all other clusters. A second SHAP analysis on these models provided an interpretable signature of each cluster—i.e., the anatomical measurements that most strongly distinguished that decision group from the remainder of the cohort.

# 3  Results

*3.1 Interpreting Model Decisions and Validating Feature Importance*

In this section we look sequentially at the four patient groupings (all BMI, lean, overweight, obese) in Figures 4-7.

We used the area under the receiver operating characteristic curve (AUC) to assess our random-forest's ability to distinguish type 2 diabetes from control patients via collections of their abdominal measurements. The mean AUCs ± standard deviation were 0.74 ± 0.02 for all BMI, 0.72 ± 0.07 for lean, 0.74 ± 0.04 for overweight, and 0.72 ± 0.06 for obese.

Our SHAP analysis of the random forest revealed both the ranking and directionality (risk vs. protective) of the top 20 anatomical measurements used in model decisions. Shared patterns across BMI groups consistently associated with type 2 diabetes included fatty muscle (lower median HU intensity), older age, increased visceral and subcutaneous fat, and a smaller or fat-laden pancreas. A fatty and/or enlarged liver was informative for type 2 diabetes in the all BMI, lean, and obese groups. Similarly, a smaller and/or fat-laden pancreas was important for predicting type 2 diabetes in those same groups. Right kidney enlargement contributed to classification toward type 2 diabetes in the all BMI, lean, and overweight groups, while fatty kidneys (left and right) were relevant for type 2 diabetes prediction in the lean group. In the overweight group, the model predicted type 2 diabetes where there was a higher density in both visceral and subcutaneous fat. In the lean group, lower BMI was paradoxically used by the model to predict type 2 diabetes, suggesting that lean individuals with type 2 diabetes may have reduced muscle mass rather than fat.

While our cross-validated random forest leveraged combinations of features to rank their importance and assign risk-versus-protective directionality, we independently validated each top feature with univariate logistic regression performed on the full dataset for every BMI subgroup (no cross-validation). The logistic regression odds-ratio direction (type 2 diabetes risk if > 1, protective if < 1) showed near-perfect concordance with the random forest SHAP findings. Moreover, after multiple comparisons correction, 18 of 20 features were significant in the all BMI group, 17 in the lean group, 14 in the overweight group, and 17 in the obese group. Because these univariate tests examined features in isolation, their general agreement with the multivariate random-forest results implies that the directionality of feature relationships with type 2 diabetes were robust across both univariate and multivariate models.

*3.2 Linking Model Decisions to Abdominal Phenotypes*

In this section we look sequentially at the four patient groupings (all BMI, lean, overweight, obese) in Figures 8-11. Previously, we looked at overall patterns that explained how our random forest decided if a patient had type 2 diabetes based on explainable measurements of their abdomen. Here we look for abdominal phenotypes or anatomical signatures in the random forest's decision space that cued the model to predict control or type 2 diabetes. We do this by looking at the top 20 abdominal measurements identified from the SHAP analysis as described in section 2.6. This enables the opportunity to discover different abdominal patterns that lead to the same model decision, which could validate or potentially uncover different mechanisms of type 2 diabetes.

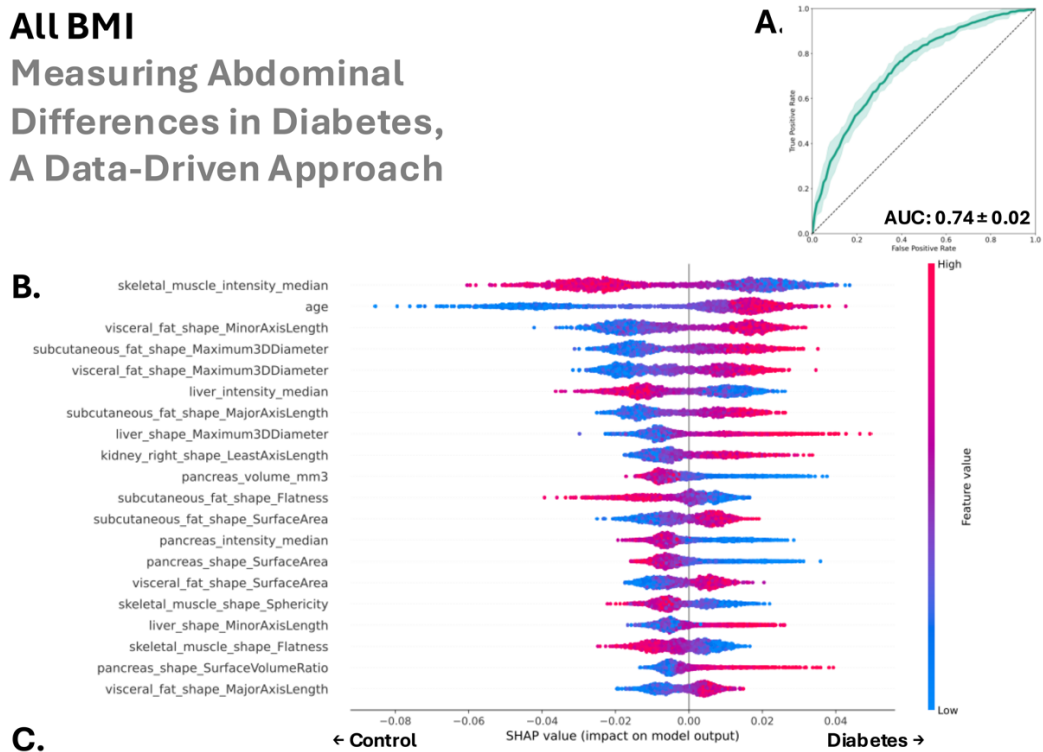

Fig. 4 Including all BMI groups, the data-driven abdominal phenotype of type 2 diabetes was characterized by reduced muscle density and sphericity, older age, greater visceral and subcutaneous fat, fatty and enlarged liver, enlarged right kidney, and a smaller fat-laden pancreas. (A) Random-forest classifier distinguishes type 2 diabetes from controls (mean AUC 0.74). (B) SHAP ranking of the top 20 features shows the magnitude and direction of their risk or protective impact. (C) Univariate logistic regression confirms the same feature associations via odds ratios, with statistical significance denoted in green. We additionally color code blue for OR < 1 and red for OR > 1, which allows these features to be compared to the right half of the SHAP plot (B)—when an OR is blue, higher values of the feature are protective of type 2 diabetes, and when an OR is red, higher values are related with type 2 diabetes risk.

# Lean
## Measuring Abdominal Differences in Diabetes, A Data-Driven Approach

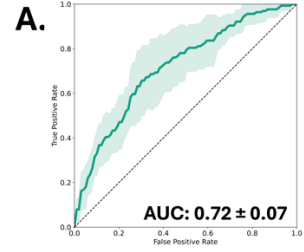

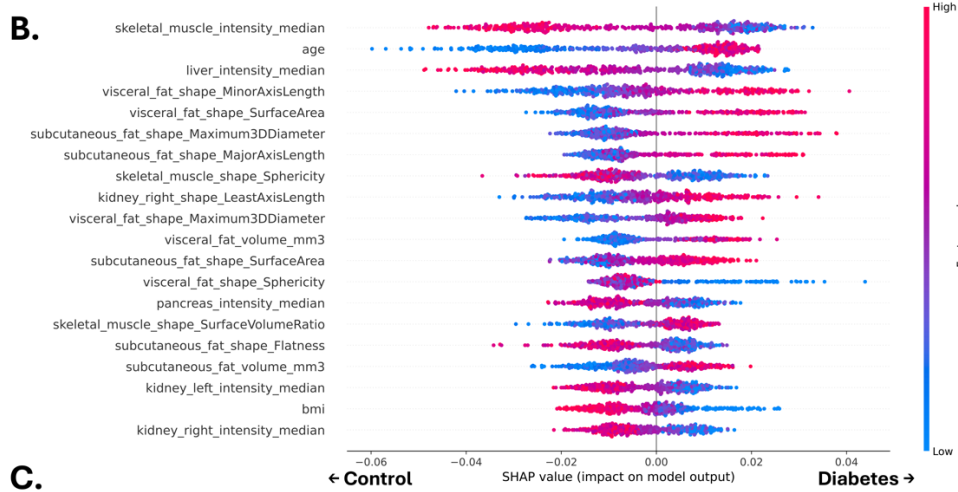

| Feature | OR | SE | 95% CI | p-value | FDR-corrected p-value |
|---|---|---|---|---|---|
| skeletal_muscle_intensity_median | 0.60 | 0.16 | 0.44–0.83 | 1.73E-03 | 3.14E-03 * |
| age | 1.81 | 0.11 | 1.46–2.25 | 6.57E-08 | 1.31E-06 * |
| liver_intensity_median | 0.45 | 0.18 | 0.31–0.64 | 9.84E-06 | 6.56E-05 * |
| visceral_fat_shape_MinorAxisLength | 1.64 | 0.13 | 1.27–2.13 | 1.61E-04 | 4.03E-04 * |
| visceral_fat_shape_SurfaceArea | 1.70 | 0.14 | 1.28–2.26 | 2.50E-04 | 5.55E-04 * |
| subcutaneous_fat_shape_Maximum3DDiameter | 1.57 | 0.11 | 1.26–1.96 | 6.63E-05 | 1.95E-04 * |
| subcutaneous_fat_shape_MajorAxisLength | 1.64 | 0.12 | 1.30–2.07 | 3.84E-05 | 1.53E-04 * |
| skeletal_muscle_shape_Sphericity | 0.83 | 0.13 | 0.64–1.08 | 1.59E-01 | 1.76E-01 |
| kidney_right_shape_LeastAxisLength | 1.49 | 0.12 | 1.18–1.88 | 7.02E-04 | 1.40E-03 * |
| visceral_fat_shape_Maximum3DDiameter | 1.91 | 0.15 | 1.43–2.57 | 1.62E-05 | 8.09E-05 * |
| visceral_fat_volume_mm3 | 1.48 | 0.13 | 1.14–1.92 | 3.36E-03 | 5.60E-03 * |
| subcutaneous_fat_shape_SurfaceArea | 1.70 | 0.13 | 1.31–2.22 | 6.81E-05 | 1.95E-04 * |
| visceral_fat_shape_Sphericity | 0.87 | 0.11 | 0.70–1.08 | 1.92E-01 | 2.03E-01 |
| pancreas_intensity_median | 0.66 | 0.17 | 0.47–0.91 | 1.23E-02 | 1.70E-02 * |
| skeletal_muscle_shape_SurfaceVolumeRatio | 1.07 | 0.15 | 0.80–1.44 | 6.48E-01 | 6.48E-01 |
| subcutaneous_fat_shape_Flatness | 0.71 | 0.13 | 0.56–0.91 | 6.85E-03 | 1.05E-02 * |
| subcutaneous_fat_volume_mm3 | 1.85 | 0.13 | 1.42–2.41 | 5.13E-06 | 5.13E-05 * |
| kidney_left_intensity_median | 0.65 | 0.17 | 0.46–0.91 | 1.28E-02 | 1.70E-02 * |
| bmi | 0.79 | 0.11 | 0.63–0.97 | 2.84E-02 | 3.34E-02 * |
| kidney_right_intensity_median | 0.69 | 0.17 | 0.50–0.96 | 2.66E-02 | 3.33E-02 * |

**Fig. 5** In the lean group (BMI < 25 kg/m$^2$), the data-driven abdominal phenotype of type 2 diabetes was characterized by reduced muscle density and sphericity, older age, fatty liver, greater visceral and subcutaneous fat, enlarged right kidney, fat-laden pancreas, fatty left and right kidneys, and counterintuitively lower BMI. (A) Random-forest classifier distinguishes type 2 diabetes from controls (mean AUC 0.72). (B) SHAP ranking of the top 20 features shows their risk or protective impact. (C) Univariate logistic regression confirms the same feature associations via odds ratios, with statistical significance denoted in green. We additionally color code blue for OR < 1 and red for OR > 1, which allows these features to be compared to the right half of the SHAP plot (B)—when an OR is blue, higher values of the feature are protective of type 2 diabetes, and when an OR is red, higher values are related with type 2 diabetes risk.

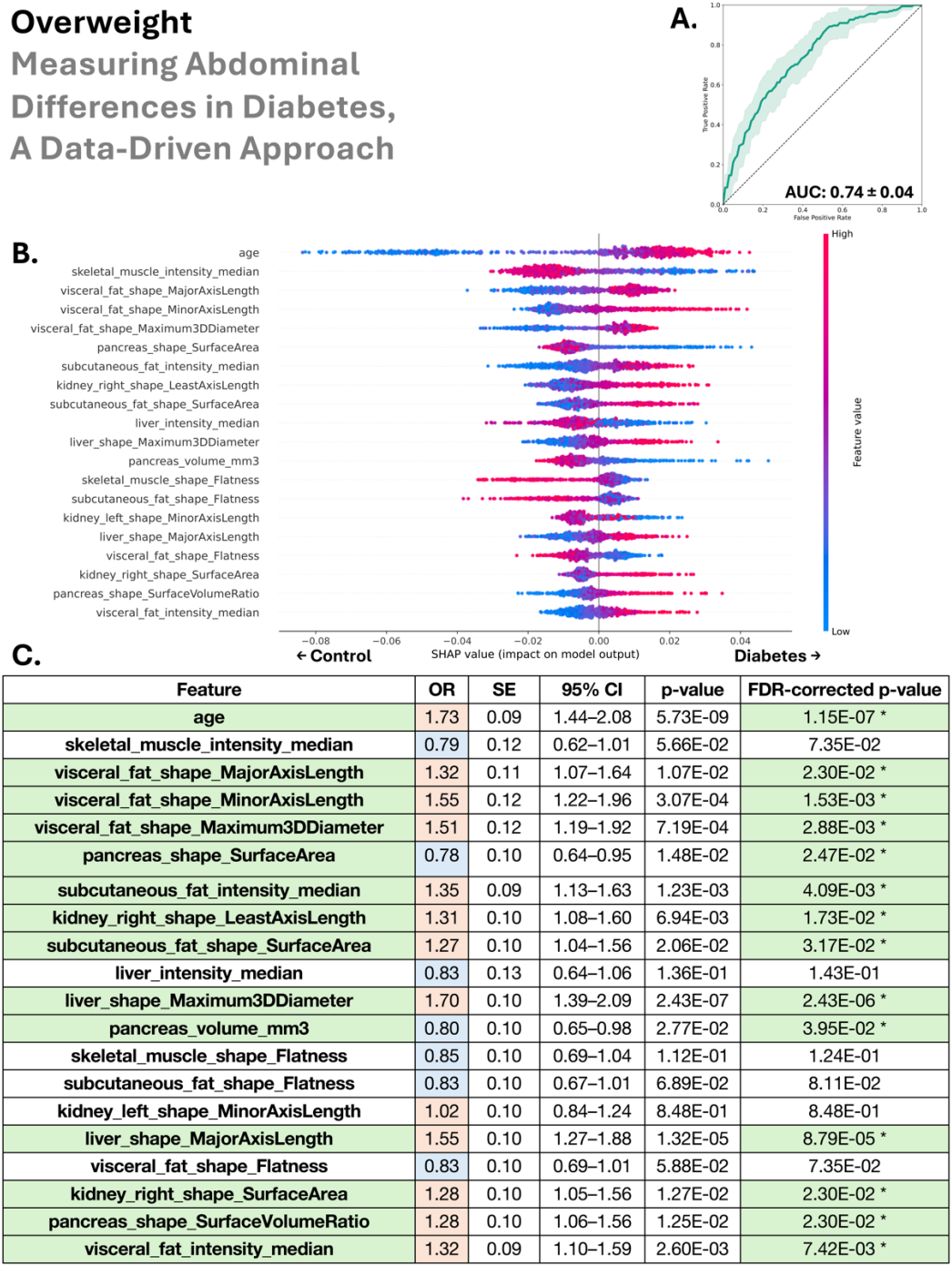

**Fig. 6** In the overweight group (25 kg/m$^2$ ≤ BMI < 30 kg/m$^2$), the data-driven abdominal phenotype of type 2 diabetes was characterized by older age, reduced muscle density, greater and denser visceral and subcutaneous fat, reduced pancreas size, and enlarged right kidneys. (A) Random-forest classifier distinguishes type 2 diabetes from controls (mean AUC 0.74). (B) SHAP ranking of the top 20 features shows their risk or protective impact. (C) Univariate logistic regression mostly agrees the same feature associations via odds ratios, with statistical significance denoted in green. We additionally color code blue for OR < 1 and red for OR > 1, which allows these features to be compared to the right half of the SHAP plot (B)—when an OR is blue, higher values of the feature are protective of type 2 diabetes, and when an OR is red, higher values are related with type 2 diabetes risk.

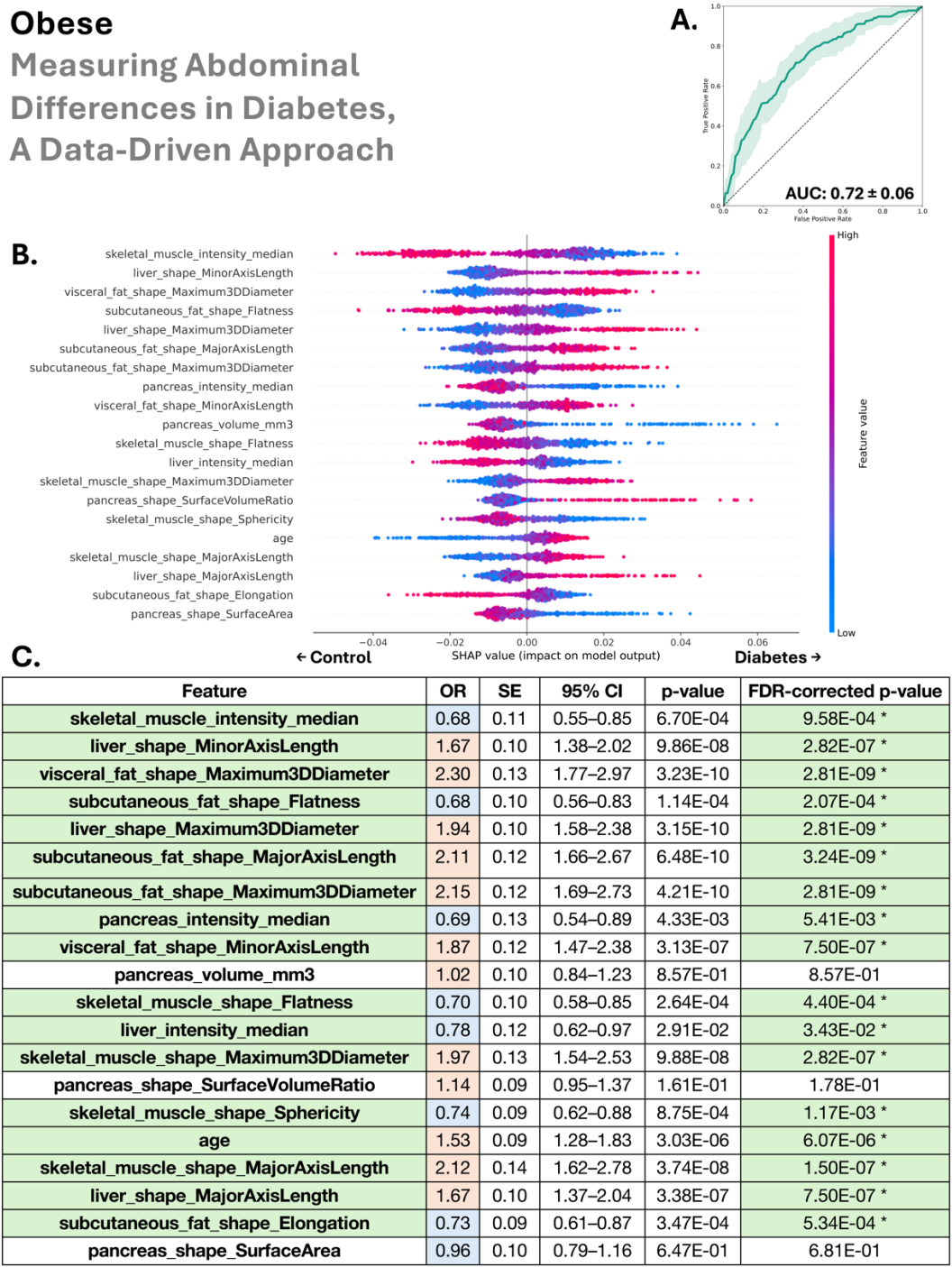

**Fig. 7** In the obese group (BMI ≥ 30 kg/m$^2$), the data-driven abdominal phenotype of type 2 diabetes was characterized by reduced muscle density and sphericity, enlarged and fatty liver, increase visceral and subcutaneous fat, sometimes smaller and fat-laden pancreas, and older age. (A) Random-forest classifier distinguishes type 2 diabetes from controls (mean AUC 0.72). (B) SHAP ranking of the top 20 features shows their risk or protective impact. (C) Univariate logistic regression mostly agrees the same feature associations via odds ratios, with statistical significance denoted in green. We additionally color code blue for OR < 1 and red for OR > 1, which allows these features to be compared to the right half of the SHAP plot (B)—when an OR is blue, higher values of the feature are protective of type 2 diabetes, and when an OR is red, higher values are related with type 2 diabetes risk.

In all the groups, as we look across the decision landscape, there is reasonable agreement between the gradient of type 2 diabetes labels and our model's predictions (Figures 8-11). Across groups, the decision space is strongly informed by age and muscle density (lower median HU) but is not driven by sex differences (all BMI, lean, overweight) or partially driven by sex differences (obese).

Separately in each of the all BMI, lean, and overweight cohorts, our method split patients into two decision clusters—one enriched for type 2 diabetes and the other for controls—with opposing anatomical signatures. In the type 2 diabetes-enriched clusters we consistently observed poorer muscle density (lower median HU) and shape, older age, and greater visceral fat. Additional type 2 diabetes-associated traits included a fatty pancreas in the all BMI and lean groups, a fatty liver in the lean and overweight groups, an elevated pancreas surface-area-to-volume ratio in the all BMI and overweight groups, and increased subcutaneous fat in the overweight group.

The obese group was automatically divided into four distinct clusters, some of which reflected sex-based anatomical patterns. One cluster was enriched for type 2 diabetes and predominantly female, characterized by a smaller, fatty pancreas, older age, reduced muscle density and mass, and a smaller liver. Another type 2 diabetes- and male-enriched cluster, with implied larger body size, showed increased visceral fat, larger liver, greater subcutaneous fat, increased skeletal muscle, and reduced muscle density. Among the control-enriched clusters, one was predominantly female and marked by reduced muscle density and size, smaller liver, lower visceral and subcutaneous fat, and a sometimes smaller pancreas with a low surface-area-to-volume ratio. The other control-enriched cluster featured better muscle density and shape, younger age, a larger non-fatty pancreas, reduced visceral fat, flatter subcutaneous fat, and a non-fatty liver.

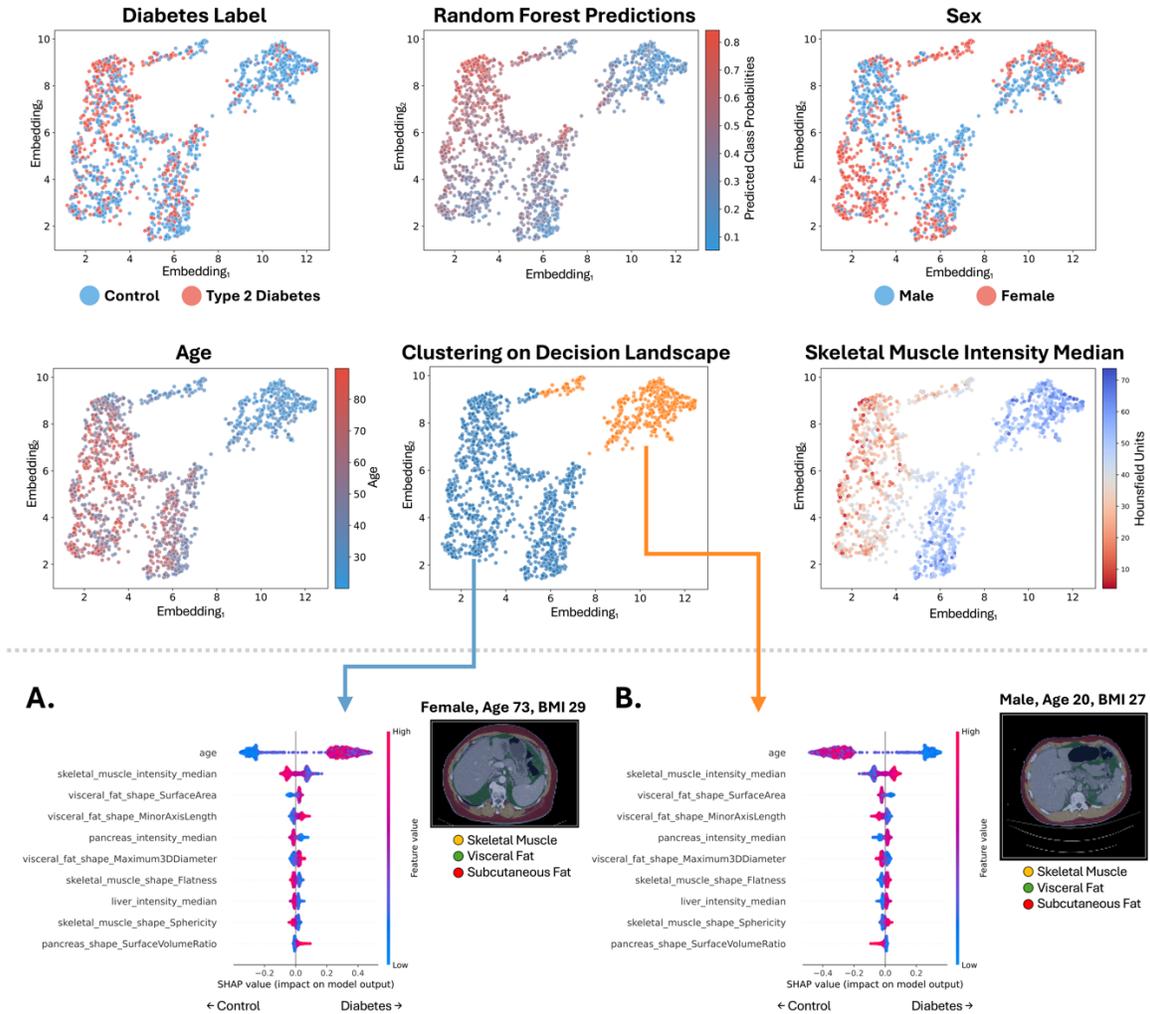

**Fig. 8** Across all BMI groups, we inspect the landscape of the random forest's decision-making approach (upper half). There is reasonable agreement between the gradient of type 2 diabetes labels and our model's predictions. The decision space is strongly informed by age and muscle density but is not driven by sex differences. Mapping decision clusters back to anatomical space, we identify a type 2 diabetes-enriched signature (A) that is characterized by older age, worsened muscle density and shape, increased visceral fat, and a fatty pancreas with high surface area to volume ratio. We see the inverse phenotype in the control patient-enriched cluster (B). We additionally visualize the abdomen from the center scan in each cluster (minimum Euclidean distance to cluster centroid).

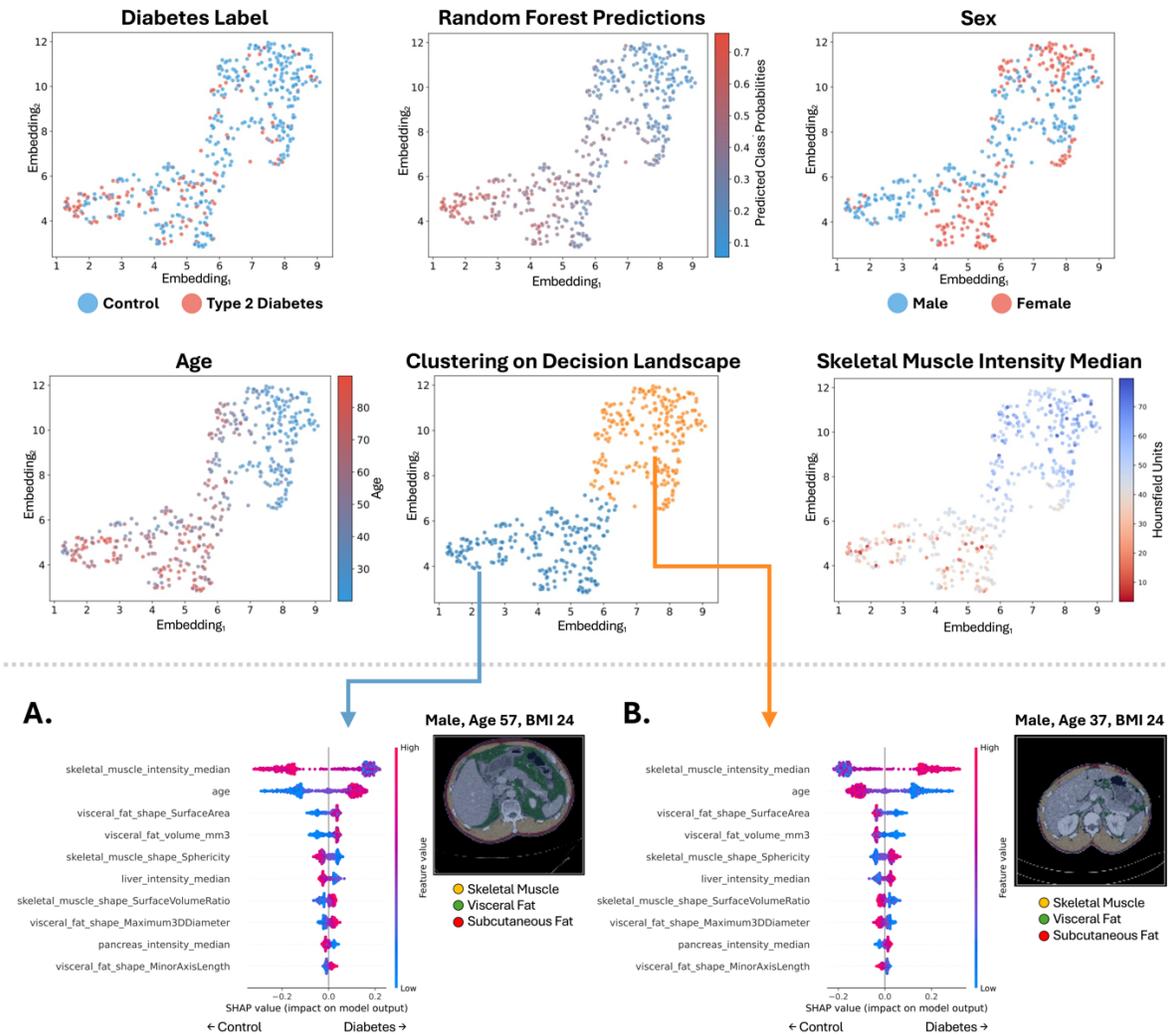

**Fig. 9** In the lean group (BMI < 25 kg/m$^2$), we inspect the landscape of the random forest's decision-making approach (upper half). There is reasonable agreement between the gradient of type 2 diabetes labels and our model's predictions. The decision space is strongly informed by age and muscle density but is not driven by sex differences. Mapping decision clusters back to anatomical space, we identify a type 2 diabetes-enriched signature (A) that is characterized by worsened muscle density and shape, older age, increased visceral fat, fatty liver, and a fatty pancreas. We see the inverse phenotype in the control patient-enriched cluster (B). We additionally visualize the abdomen from the center scan in each cluster (minimum Euclidean distance to cluster centroid).

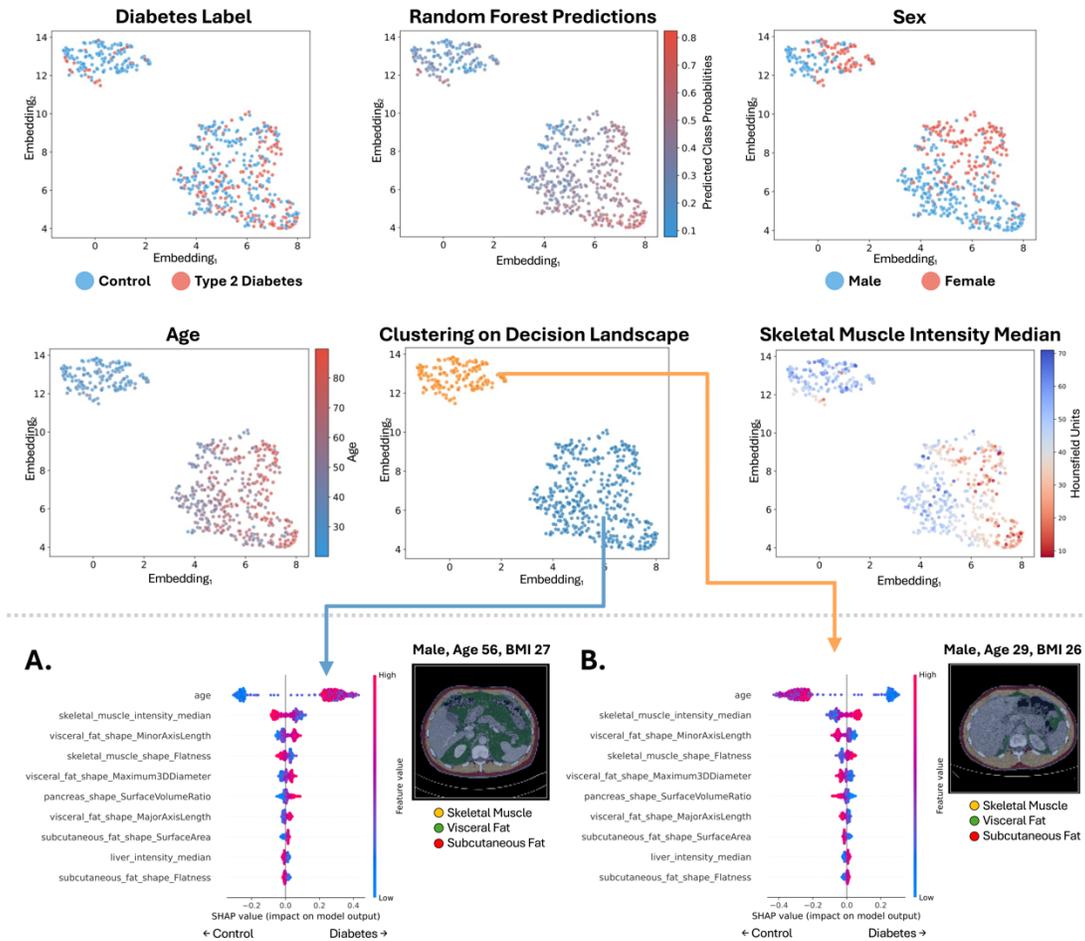

**Fig. 10** In the overweight group (25 kg/m2 ≤ BMI < 30 kg/m$^2$), we inspect the landscape of the random forest's decision-making approach (upper half). There is reasonable agreement between the gradient of type 2 diabetes labels and our model's predictions. The decision space is strongly informed by age and muscle density but is not driven by sex differences. Mapping decision clusters back to anatomical space, we identify a type 2 diabetes-enriched signature (A) that is characterized by older age, worsened muscle density and shape, increased visceral fat, increased subcutaneous fat, increased pancreas surface area to volume ratio, and a fatty liver. We see the inverse phenotype in the control patient-enriched cluster (B). We additionally visualize the abdomen from the center scan in each cluster (minimum Euclidean distance to cluster centroid).

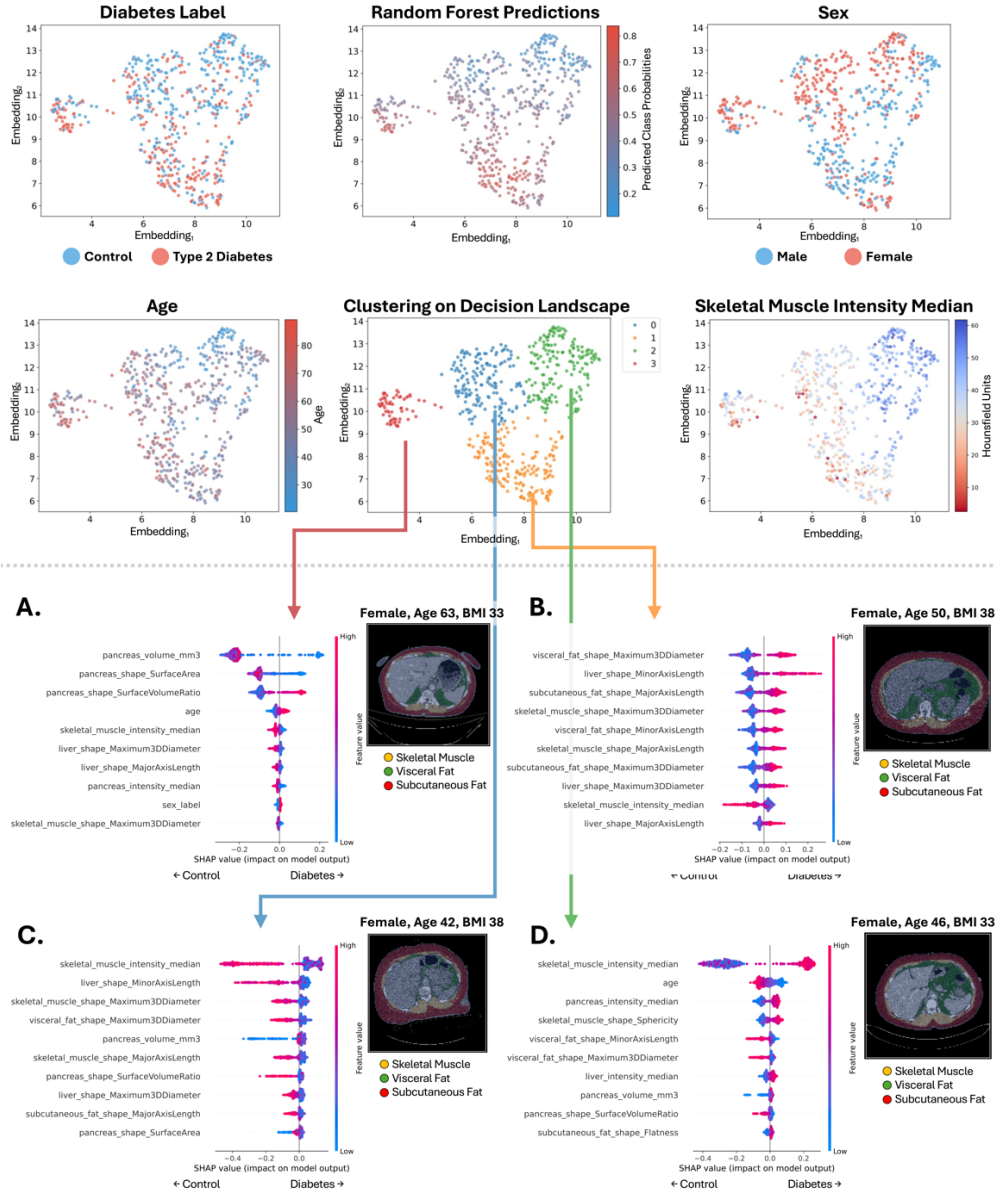

**Fig. 11** In the obese group (BMI > 30 kg/m$^2$), we inspect the landscape of the random forest's decision-making approach (upper half). There is reasonable agreement between the gradient of type 2 diabetes labels and our model's predictions. The decision space is strongly informed by age and muscle density and partially driven by sex differences. Mapping decision clusters back to anatomical space, we identify a type 2 diabetes- and female-enriched signature (A) that is characterized by a smaller fatty pancreas, older age, reduced muscle density and mass, and a smaller liver. In B, we identify a type 2 diabetes- and male/large-body-enriched signature that is characterized by increased visceral fat, increased liver size, increased subcutaneous fat, increased skeletal muscle, and reduced muscle density. In C, we identify a control- and female-enriched signature that is characterized by worsened muscle density and size, smaller liver, reduced visceral and subcutaneous fat, and a sometimes smaller pancreas with a low surface area to volume ratio. In D, we identify a control-enriched cluster that is characterized by better muscle density and shape, lower age, a larger non-fatty pancreas, reduced visceral fat and flatter subcutaneous fat, and a non-fatty liver.

# 4   Discussion

By leveraging explainable machine learning, we uncovered multi-feature anatomical profiles associated with type 2 diabetes and demonstrated that exploring the model's decision space enables automatic identification of clinically meaningful phenotypes directly within the anatomical measurement space. These findings support a shift toward comprehensive, data-driven characterization of type 2 diabetes risk and phenotype signatures.

The type 2 diabetes classification random forests were not exceptional diagnostic models (mean AUCs 0.72–0.74); an AUC of 0.5 is akin to random chance and an AUC of 1.0 is a perfect classifier. However, these random forests had reasonable AUCs for biological pattern discovery. In the literature, AUCs near 0.7 are acceptable in the context of image-based models of metabolic disease[53,54]. Our AUC range was not surprising since we were classifying patients based only on measurements of their abdomen, sex, age, BMI, and contrast flag, but not including more powerful clinical measurements of blood glucose or medications.

Our data-driven signature of type 2 diabetes agreed with findings in the literature specifying that, when compared to normal, there is more subcutaneous fat, more visceral fat, and more fatty lean mass[9,10,11,12,13,14,15,16,17]. Additionally, our type 2 diabetes signature shows concordance with the field on fatty liver, fatty pancreas, smaller pancreas, and sometimes fatty kidneys[19,20,21,21,22,23,24,25–27,15,23,28]. Our signature highlights the importance of fatty skeletal muscle in predicting type 2 diabetes across the board, which directly contrasts some of the bioelectrical impedance analysis literature that states lean mass is not protective of type 2 diabetes[30,31].

Our findings suggest that image-based phenotyping may support type 2 diabetes risk profiling from secondary use of CT scans acquired for other clinical purposes. Our study focused on data from one site. Future work should explore the generalizability of such type 2 diabetes risk profiling approaches across imaging sites, ultimately evaluating their potential for broader clinical utility.

## 5 Conclusions

Our study identified type 2 diabetes patterns in the abdomen that were largely consistent with the literature. Separately for lean, overweight, and obese groups, fatty skeletal muscle emerged as either the most or second most important predictor, underscoring its central role in type 2 diabetes. Moreover, we found that the abdominal patterns of type 2 diabetes were largely the same across the separate lean, overweight, and obese groups, suggesting shared anatomical signatures of type 2 diabetes despite BMI class. While we focused on type 2 diabetes in this work, our phenotyping methodology could be applied for pattern discovery in other disease domains.

*Disclosures*

The authors declare that there are no financial interests, commercial affiliations, or other potential conflicts of interest that could have influenced the objectivity of this research or the writing of this paper.

*Data Availability*

The data that support the findings of this article are not publicly available.

*Acknowledgments*

This work was supported by Integrated Training in Engineering and Diabetes, grant number T32 DK101003 and MSTP T32: NIH NIGMS T32GM007347. This work used REDCap and VCTRS resources, which are supported by grant UL1 TR000445 from National Center for Advancing Translational Sciences, National Institutes of Health (NIH). These studies benefited from the infrastructure of the Vanderbilt Diabetes Research and Training Center (DK 020593), Division of Diabetes, Endocrinology, and Metabolic Diseases, and the Vanderbilt University Institute of Imaging Science Center for Human Imaging (1 S10OD021771 01). This work was supported by NSF career 1452485 and NSF 2040462. This work was supported by the Alzheimer's Disease Sequencing Project Phenotype Harmonization Consortium (ADSP-PHC) that is funded by NIA (U24 AG074855, U01 AG068057 and R01 AG059716). This work was conducted in part using the resources of the Advanced Computing Center for Research and Education at Vanderbilt University, Nashville, TN. The Vanderbilt Institute for Clinical and Translational Research (VICTR) is funded by the National Center for Advancing Translational Sciences (NCATS) Clinical Translational Science Award (CTSA) Program, Award Number 5UL1TR002243-03. The content is solely the responsibility of the authors and does not necessarily represent the official views of the NIH. This work was supported by DoD grant HT94252410563. We extend gratitude to NVIDIA for their support by means of the NVIDIA hardware grant. This research was funded by the National Cancer Institute (NCI) grant R01 CA253923-04, R01 CA 253923-04S1. Financial support was graciously provided by the National Institutes of Health (DK129979 and HD11556) and Breakthrough T1D (formerly JDRF) (1-INO-2023-1340-A-N). We gratefully acknowledge philanthropic support from Thomas J. and Karen K. Gentry. The study sponsors were not involved in the design of the study and did not impose any restrictions regarding the publication of the report. We have used AI as a tool in the creation of this content, however, the foundational ideas, underlying concepts, and original gist stem directly from the personal insights, creativity, and intellectual effort of the author(s). The use of generative AI serves to enhance and support the author's original contributions by assisting in the ideation, drafting, and refinement processes. All AI-assisted content has been carefully reviewed, edited, and approved by the author(s) to ensure it aligns with the intended message, values, and creativity of the work.